\documentclass{article}
\usepackage{spconf,amsmath,graphicx, caption, subcaption, amssymb, algorithm, mathrsfs}

\title{Sparse And Low Rank Decomposition Based Batch Image Alignment for Speckle Reduction of Retinal OCT Images}
\name{Ahmadreza Baghaie\textsuperscript{1}, Roshan M. D'souza\textsuperscript{2}, Zeyun Yu\textsuperscript{3}}
\address{Depts. of \textsuperscript{1}Electrical Engineering, \textsuperscript{2}Mechanical Engineering and \textsuperscript{3} Computer Science,\\ University of Wisconsin-Milwaukee, WI, USA}
\begin{document}
%\ninept
%
\maketitle
\begin{abstract}
Optical Coherence Tomography (OCT) is an emerging technique in the field of biomedical imaging, with applications in ophthalmology, dermatology, coronary imaging etc. Due to the underlying  physics, OCT images usually suffer from a granular pattern, called speckle noise, which restricts the process of interpretation. Here, a sparse and low rank decomposition based method is used for speckle reduction in retinal OCT images. This technique works on input data that consists of several B-scans of the same location. The next step is the batch alignment of the images using a sparse and low-rank decomposition based technique. Finally the denoised image is created by median filtering of the low-rank component of the processed data. Simultaneous decomposition and alignment of the images result in better performance in comparison to simple registration-based methods that are used in the literature for noise reduction of OCT images. 
\end{abstract}
\section{Introduction}
\label{sec:intro}

Optical Coherence Tomography (OCT) is a powerful imaging system for non-invasive acquisition of 3D volumetric images of tissues. From its emerge in the early 1990's \cite{huang1991optical} until now, many improvements have been achieved regarding the OCT imaging system. Nowadays, taking $\mu m$-level resolution volume images of the tissues are very common, especially in ophthalmology. Therefore the need for specialized OCT image analysis techniques is of high interest, making this field by far one of the most attracting areas in biomedical imaging. 

One important task in OCT image processing is the removal of speckle noise. Speckle is a fundamental property of the signals and images acquired by narrow-band detection systems like Synthetic-Aperture Radar (SAR), ultrasound and OCT. Not only the optical properties of the system, but also the motion of the subject to be imaged, size and temporal coherence of the light source, multiple scattering, phase deviation of the beam and aperture of the detector can affect the speckle \cite{schmitt1999speckle}. Fig. 1(a) shows a sample retinal OCT image, highly degraded by speckle noise.

Speckle is considered to be multiplicative noise, in contrast to the additive Gaussian noise. Due to limited dynamic range of displays, OCT signals are usually compressed by a logarithmic operator applied to the intensity information which converts the multiplicative speckle noise to additive noise \cite{salinas2007comparison}. OCT noise reduction techniques can be divided into two major classes: 1) methods of noise reduction during the acquisition time and 2) post-processing techniques. In the first class, which is usually referred to as "compounding techniques", multiple uncorrelated recordings are averaged. These include spatial compounding \cite{avanaki2013spatial}, angular compounding \cite{schmitt1997array}, polarization compounding \cite{kobayashi1991polarization} and frequency compounding \cite{pircher2003speckle}. There are two major classes of post-processing techniques  for speckle noise reduction: anisotropic diffusion-based techniques \cite{salinas2007comparison} and multi-scale/multi-resolution geometric representation techniques \cite{pizurica2008multiresolution}. Use of compressive sensing and sparse representation have also been explored in the past few years \cite{fang2013fast}. For a more complete review on the different image analysis techniques in OCT image processing, including noise reduction, the reader is referred to \cite{baghaie2014state} and references therein. 

Post-processing averaging/median filtering is also an interesting method for speckle reduction. Usually in such techniques, multiple B-scans of the same location are acquired and then the average/median is taken. The main assumption here is having uncorrelated speckle between different recordings which is mostly satisfied. The misalignment between the different B-scans is usually compensated with a parametric image registration technique, such as translation based registration, rigid registration or affine registration. Theoretically, having $ N $ B-scans, the Signal-to-Noise-Ratio (SNR) can be improved by a factor of  $ \sqrt{N} $. 

\begin{figure*}

\begin{subfigure}[b]{0.5\textwidth}
\centering
\includegraphics[scale=.25]{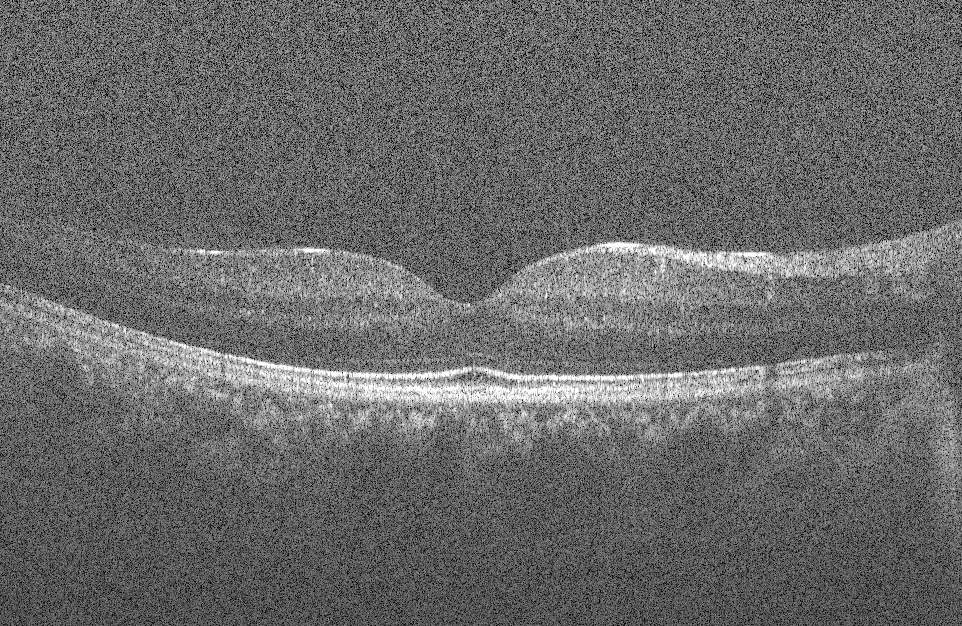}
\caption{}
\end{subfigure}\
\begin{subfigure}[b]{0.5\textwidth}
\centering
\includegraphics[scale=.25]{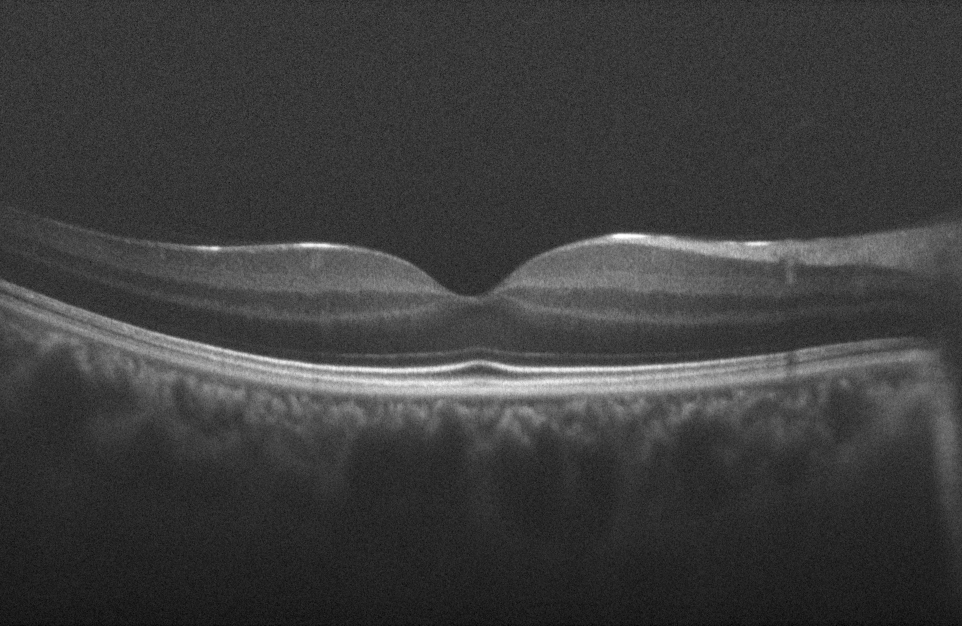}
\caption{}
\end{subfigure}

\caption{(a) Sample retinal OCT image degraded by speckle noise, (b) Final result of the proposed method using 50 misaligned noisy retinal OCT images}
\end{figure*}

In \cite{jorgensen2007enhancing} a dynamic programming based method is used for compensation of the translational movements between several B-scans and reducing the speckle noise. But as it is obvious, translation is not the only possible movement that can happen between different B-scans. A hierarchical model-based motion estimation scheme based on an affine-motion model is used in \cite{alonso2011speckle} for registering multiple B-scans to be used for speckle reduction.

In this paper, another technique for registration-based speckle reduction is proposed. This technique utilizes sparse and low rank decomposition to separate between image features and noise components in each B-scan, while aligning them iteratively. Using this technique, sub-pixel accuracy can be achieved for the alignment process which can further improve the SNR and Contrast-to-Noise-Ratio (CNR) in the final denoised result. The paper is organized as follows: Section 2 contains detailed explanation of the sparse and low rank decomposition based batch image alignment technique. In  Section 3 the results of this method is presented and compared with translation and rigid registration methods in terms of SNR and CNR. Section 4 concludes the paper with further pointers regarding future investigations.

\section{Methods}
\subsection{Robust Principal Component Analysis (RPCA)}

Given a large data matrix $D\in \mathbb{R}^{m\times n}$, the RPCA process divides $D$ into two components: $D=L+S$, with $L$ being the low rank component  and $S$ being the sparse component \cite{candes2011robust}. In an $ l_2 $ sense, this is the classic Principal Component Analysis (PCA). Even though this technique is widely used in the literature, the inherent sensitivity to outliers makes it less useful in modern applications. This can be remedied by minimizing $||L||_*+\lambda ||S||_1$ s.t $L+S=D$ which is proven to have an exact recovery under broad conditions \cite{candes2011robust}. 

This concept has been widely used in different branches of computer vision and image processing such as video surveillance, shadow or specularity removal in face recognition, video repairing etc. In \cite{luan2013application} this technique is used for single OCT image noise reduction. For post-processing averaging/median filtering, multiple B-scans of the same location are acquired and are used for noise reduction. As mentioned before, here the main issue is regarding the misalignment between the B-scans, as well as the differences in the displayed patterns due to eye movement. This requires image alignment prior to averaging/median filtering. One technique is to pre-register the noisy images and use them as inputs for the next stage. Considering the high amount of noise degrading the images, this can cause erroneous alignment. A better way is to combine the image registration task with low rank/sparse decomposition of the data which reduces the effect of noise in the process of registration. Another advantage is its ability to detect the underlying low-rank pattern which results in elimination of retinal features that are not present in all of the slices and only appear due to eye movement. Here, we follow the work of \cite{peng2012rasl} for simultaneous alignment and decomposition of the retinal OCT images.

\subsection{Robust Alignment by Sparse and Low-Rank Decomposition (RASL)}

Assuming a set of $n$ B-scans, the data matrix $D$ can be created by stacking the vectorized images as columns of the matrix. Having completely aligned images, it is expected that $D$ is low-rank, with the possibility of having sparse additional noise: $ D = L+S $. In the case of OCT images considered here, additional optimization is needed in order to compensate the misalignment. This is done by assuming a set of parametric transformations, $ \tau $, applied to the images. In this formulation we have:
 \begin{equation}
\textrm{min}_{L, S, \tau} rank(L)+\lambda ||S||_0 \quad \textrm{s.t.} \quad D\circ \tau=L+S 
\end{equation}

This is a non-convex and NP-hard problem to solve due to the need for minimizing the rank and the $l_0$ norm. Convex relaxation of the problem as elaborated in \cite{candes2011robust} results in:
\begin{equation}
\textrm{min}_{L, S, \tau} ||L||_*+\lambda ||S||_1 \quad \textrm{s.t.} \quad D\circ \tau=L+S 
\end{equation}
where $||.||_*$ is the nuclear norm  (sum of singular values) and $||.||_1$ is the $l_1$ norm.

Another difficulty arises from the non-linearity of the constraint $D\circ \tau=L+S$ . This can be solved assuming minimal changes in $\tau$ in each iteration and linearizing around the current estimate of $\tau$. Therefore $ D\circ(\tau+\Delta \tau) \approx D\circ \tau+ \sum _{i=1}^n J_i \Delta \tau \varepsilon _i \varepsilon _i ^ T $ where $J_i$ is the Jacobian of the $i$th image with respect to the transformation parameters and ${\varepsilon_i}$ is the standard basis for $\mathbb{R}^n$. As mentioned in \cite{peng2012rasl} this linearization only holds for small misalignment between the images in the batch. Starting from an initial set of transformation, here the identical transformation, and setting rigid transformation as the desired transformation, at each iteration this linearized convex optimization problem is solved using the normalized images to avoid the trivial solutions until reaching convergence. Algorithm 1 summarizes the process.

\begin{algorithm}
\caption{Sparse and Low-rank Based Alignment}
\textbf{Inputs}: input images, initial transformation set, $\lambda>0$\\
\textbf{WHILE} not converged \textbf{DO}

\quad \textbf{Step 1:} compute the Jacobian w.r.t transformations: $J_i$

\quad \textbf{Step 2:} warp and normalize the images: $D\circ \tau$  

\quad \textbf{Step 3 (inner loop):} solve the linearizied convex optimization problem:
\begin{multline*}
 (L^*, S^*, \Delta \tau^*) \leftarrow \textrm{arg min}_{L, S, \Delta \tau} ||L||_*+\lambda ||S||_1 \\
 \textrm{s.t.} \quad D\circ \tau+ \sum _{i=1}^n J_i \Delta \tau \varepsilon_i \varepsilon _i ^ T=L+S  
\end{multline*}

\quad \textbf{Step 4:} update transformation: $\tau\leftarrow\tau+\Delta\tau$

\textbf{END WHILE}

\textbf{OUTPUT:} solution $L^*$, $S^*$ and $\tau^*$ to problem (2). 

\end{algorithm}

The main computational cost of the Algorithm 1 is in the third step: solving the linearized convex optimization problem. Considering the possibility of having millions of variables, having a scalable solution is of high importance. Augmented Lagrange Multiplier (ALM) \cite{lin2010augmented} has been proven to have reliable results for such optimization. Defining $h(L, S, \Delta \tau)= D\circ(\tau+\Delta \tau) \approx D\circ \tau+ \sum _{i=1}^n J_i \Delta \tau \varepsilon _i \varepsilon _i ^ T -L-S$, the augmented Lagrangian function to be optimized is defined as:
\begin{multline}
\mathcal{L}_\mu(L,S,\Delta \tau, Y)=||L||_*+\lambda ||E||_1\\+<Y, h(L, S, \Delta \tau)> +\frac{\mu}{2}||h(L, S, \Delta \tau)||_F^2
\end{multline}
where $ Y $ is the Lagrange multiplier matrix, $\mu$ is a positive scalar and $||.||_F$ is the Frobenius norm. Choosing an appropriate $Y$ and large enough $\mu$, the augmented Lagrangian function has the same minimizer as the  original constrained optimization problem. For further explanations regarding the optimization process the reader is referred to \cite{peng2012rasl, lin2010augmented}.

The final result of the algorithm is the well-aligned stack of images, decomposed into low-rank data set containing image information and sparse component consisting of speckle noise. As investigated in \cite{avanaki2013spatial}, spatial compounding works best with use of median filtering rather than averaging. Here, the final image is created by pixel-wise median filtering of the final low-rank component of the data. 

\section{Results and Discussion}

For assessing the performance of the proposed method, several metrics are considered. Considering 6 regions of interest (ROI) in the final results, one only containing background noise and the rest containing image features, the metrics can be defined as follows:
\begin{equation}
\begin{split}
SNR_m=20\times log(\frac{\mu _m}{\sigma _b}) \\
 CNR_m=\frac{\mu _m -\mu _b}{\sqrt{\sigma ^2_m+\sigma ^2_b}} \\
\end{split}
\end{equation}
where $ \mu_b $ and $ \sigma_b $ are the mean and standard deviation of the background noise and $ \mu_m $ and $ \sigma_m $ are the mean and standard deviation of the $ m $-th  ROI containing image features. The average of these metrics are considered here for comparison.

Different numbers of images of the human retina in the central foveal region are considered for assessing the performance of the proposed algorithm. Fig. 1(a) shows a sample image of the dataset that is used here. As for other registration based methods, translation and rigid registration techniques available in ImageJ \cite{schneider2012671} software package are considered for comparison, since they are widely used in the literature and give reasonable performance. 

Fig. 2 represents the improvement in the average SNR of the final image for different techniques, while Fig. 3 shows the improvement achieved in average CNR for different number of input images. Fig.1 (b) shows the final result of the proposed algorithm for speckle noise reduction using 50 misaligned noisy OCT input images. 

One critical step in post-processing averaging/median filtering of the OCT images for noise reduction is the pre-selection of the set of images to be registered and averaged. This is due to the presence of $\mu m$-level features in the high resolution OCT images. During the imaging session, movement of the eye in 3 dimensions causes these features to appear/disappear between consecutive B-scans. In other words, fine features from very close locations come to focus, while some other features will go out of focus. This makes the pre-selection a necessary step at the beginning of the process. Using RPCA, this can be eliminated. As mentioned before, RPCA tries to divide the input set of data, here the stack of vectorized B-scans, into two components: one low-rank and one sparse. In this scenario, the algorithm looks for similar patterns in the data that are shared between different B-scans to extract the low-rank component while eliminating different patterns as being noise. In other words, without the need for pre-selection, the RPCA chooses the most common features as the ground truth and neglects the features that only appear in few B-scans and considers them as noise. Even though this can be achieved using simple averaging too, given enough number of images, still the main drawback is that naive averaging is indecisive about the common/uncommon features to be averaged causing it to have more blurred features. This is because the uncommon features are diffused to the rest of the data. The same analogy can be applied for elimination of blood vessel shadows between different B-scans. Fig. 4 displays close-ups of the original and filtered version of the input images using different techniques for comparison. 

\begin{figure}
\centering
\includegraphics[scale=.40]{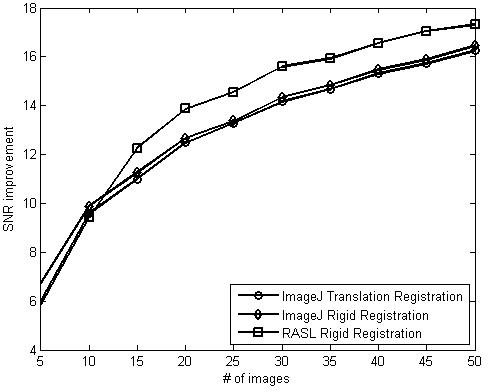}
\caption{SNR improvement for different number of input images (5-50)}
\end{figure}

\begin{figure}
\centering
\includegraphics[scale=.40]{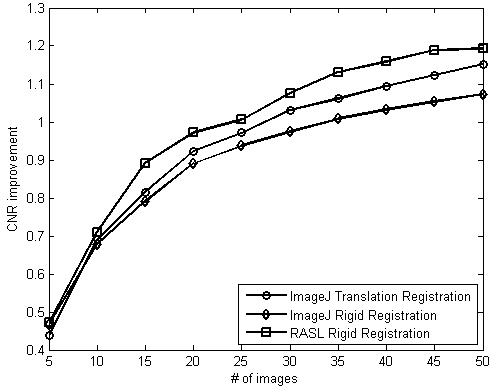}
\caption{CNR improvement for different number of input images (5-50)}
\end{figure}

\begin{figure}

\begin{subfigure}[b]{0.1\textwidth}
\centering
\includegraphics[scale=.45]{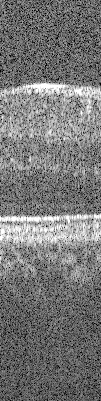}
\caption{}
\end{subfigure}
\begin{subfigure}[b]{0.1\textwidth}
\centering
\includegraphics[scale=.45]{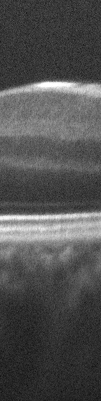}
\caption{}
\end{subfigure}
\begin{subfigure}[b]{0.1\textwidth}
\centering
\includegraphics[scale=.45]{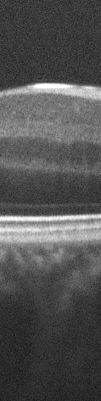}
\caption{}
\end{subfigure}
\begin{subfigure}[b]{0.1\textwidth}
\centering
\includegraphics[scale=.45]{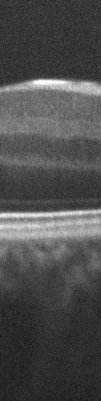}
\caption{}
\end{subfigure}

\caption{A portion of one of the input images (a) and the results of ImageJ translation (b), ImageJ rigid (c) and RASL rigid registration based methods.}
\end{figure}

\section{Conclusion}

In this paper, a new application of the sparse and low rank decomposition based batch image alignment in noise reduction of OCT images is introduced. Having a stack of misaligned, mostly due to eye movements, and noisy retinal OCT images, the process of alignment is done by decomposition of the vectorized image data into low rank and sparse components at each iteration to ensure better final alignment and noise/signal separation. Using SNR and CNR as metrics, the performance of the method is compared with some other registration based techniques for speckle noise reduction. Our approach gives better performance when bench marked against other techniques with respect to measures such as SNR and CNR while incurring larger computational cost. Further investigations are needed for speeding up the process using GPU implementations. Also, from an algorithmic point of view, newer techniques have been proposed in the literature for sparse and low rank decomposition, which will be considered in future research.

\section{Acknowledgment }
This work was partially supported by NIH P30EY001931. The authors would like to thank Dr. Joseph Carroll from Advanced Ocular Imaging Program (AOIP), Medical College of Wisconsin (MCW), Milwaukee, WI for providing the data and insight.

\bibliographystyle{IEEEbib}
\bibliography{refs.bib}

\end{document}